\documentclass[11pt,a4paper]{article}
\usepackage[hyperref]{emnlp2020}
\usepackage{times}
\usepackage{latexsym}

\usepackage{microtype}

\aclfinalcopy % Uncomment this line for the final submission
\usepackage{booktabs}
\usepackage{amsmath}
\usepackage{graphicx}
\usepackage{multirow}
\usepackage{array}
\usepackage{bm}
\usepackage{enumitem}
\newcommand{\PreserveBackslash}[1]{\let\temp=\\#1\let\\=\temp}
\newcolumntype{C}[1]{>{\PreserveBackslash\centering}p{#1}}
\newcolumntype{R}[1]{>{\PreserveBackslash\raggedleft}p{#1}}
\newcolumntype{L}[1]{>{\PreserveBackslash\raggedright}p{#1}}

\newcommand{\tabincell}[2]{\begin{tabular}{@{}#1@{}}#2\end{tabular}}

\title{From Bag of Sentences to Document: Distantly Supervised Relation Extraction via Machine Reading Comprehension}

\author{
  Lingyong Yan\textsuperscript{\rm 1,3},
  Xianpei Han\textsuperscript{\rm 1,2,*},
  Le Sun\textsuperscript{\rm 1,2},
  Fangchao Liu\textsuperscript{\rm 1,3},
  Ning Bian\textsuperscript{\rm 1,3}\\
  \textsuperscript{\rm 1}Chinese Information Processing Laboratory \ 
  \textsuperscript{\rm 2}State Key Laboratory of Computer Science\\
  Institute of Software, Chinese Academy of Sciences, Beijing, China\\
  \textsuperscript{\rm 3}University of Chinese Academy of Sciences, Beijing, China\\
  \{lingyong2014, xianpei, sunle, fangchao2017, bianning2019\}@iscas.ac.cn
}
\date{}

\begin{document}
\maketitle
\begin{abstract}
  Distant supervision (DS) is a promising approach for relation extraction but often suffers from the noisy label problem. Traditional DS methods usually represent an entity pair as a bag of sentences and denoise labels using multi-instance learning techniques.
  The bag-based paradigm, however, fails to leverage the inter-sentence-level and the entity-level evidence for relation extraction, and their denoising algorithms are often specialized and complicated.
  In this paper, we propose a new DS paradigm--document-based distant supervision, which models relation extraction as a document-based machine reading comprehension (MRC) task.
  By re-organizing all sentences about an entity as a document and extracting relations via querying the document with relation-specific questions, the document-based DS paradigm can simultaneously encode and exploit all sentence-level, inter-sentence-level, and entity-level evidence.
  Furthermore,  we design a new loss function--DSLoss (distant supervision loss), which can effectively train MRC models using only $\langle$document, question, answer$\rangle$ tuples, therefore noisy label problem can be inherently resolved.
  Experiments show that our method achieves new state-of-the-art DS performance.
  \end{abstract}
  
  \section{Introduction}
  Relation extraction (RE) is a fundamental task of natural language processing (NLP), which aims to identify relations between entities in the raw text.
  Due to the lack of large-scale manually labeled data, distant supervision\cite{mintz_distant_2009,hoffmann_knowledgebased_2011} is a promising approach for relation extraction, which heuristically generates labeled data by aligning relational tuples (e.g., $\langle$\texttt{Obama}, \texttt{birthplace}, \texttt{United States}$\rangle$) from knowledge bases (KBs) with sentences in the raw text (e.g., ``Obama was born in the United States.'').
  Unfortunately, DS approaches often suffer from the noisy label problem \cite{riedel_modeling_2010,hoffmann_knowledgebased_2011,zeng_distant_2015}, i.e., not all distantly labeled sentences express the relationship between the head entity and the tail entity in a KB.
  
  \begin{figure}[!tbp]
    \setlength{\belowcaptionskip}{-1em}
    \centering  
    \includegraphics[width=\columnwidth]{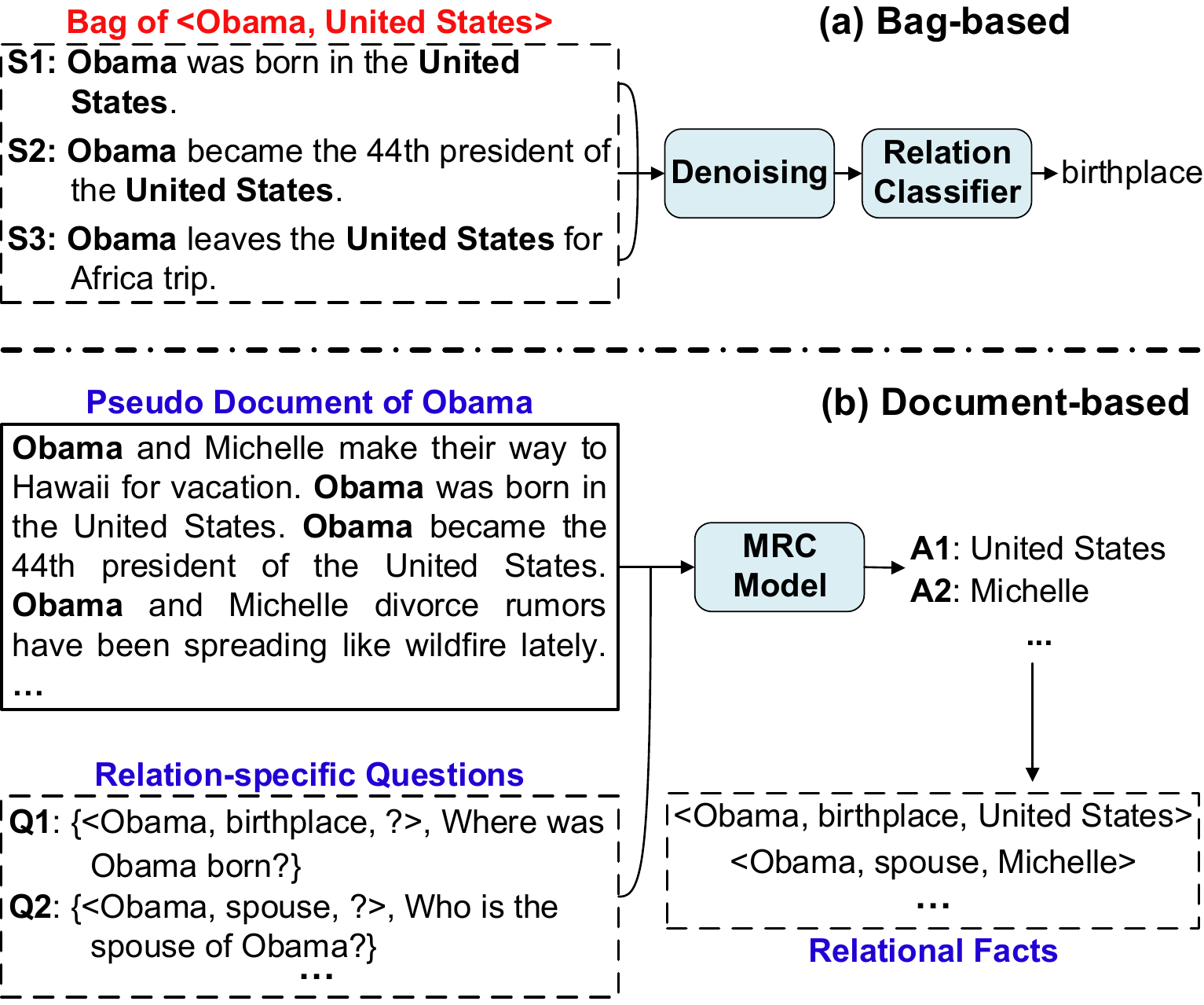}
   \caption{Illustration of the bag-based DS paradigm (a), and the proposed document-based DS paradigm (b).}
    \label{fig:fig1}
  \end{figure}
  
  To address the noisy label problem, most current DS methods \cite{riedel_modeling_2010,zeng_distant_2015,lin_neural_2016, yuan_crossrelation_2018, ye_distant_2019} employ multi-instance learning techniques, which: 
  1) first represent all sentences containing the same entity pair as a bag; 
  2) then designing denoising algorithms to distinguish relevant sentences from noisy sentences;
  3) finally predict the relation label of a bag by aggregating evidence from relevant sentences.
  For example, in Figure 1(a), the bag of $\langle$\texttt{Obama}, \texttt{United States}$\rangle$ contains three sentences, and only S1 expresses birthplace relation.
  This paper refers to this paradigm as the \textbf{\emph{bag-based paradigm}}.
  
  Despite its promising performance, the bag-based DS paradigm has two main drawbacks:
  1) Its denoising algorithm is often complicated and depends on some specialized assumptions, which may fail in open situations.
  For example, the widely used expressed-at-least-one assumption \cite{riedel_modeling_2010} often fails when all sentences in a bag are noisy \cite{ye_distant_2019}.
  2) The bag-based paradigm models different sentences independently, cannot leverage inter-sentence-level and entity-level evidence effectively.
  For example, in Figure 1(a), although S2 does not directly express the birthplace between Obama and the United States, it can still provide helpful confidence evidence for their birthplace relation.
  
  In this paper, we propose a new DS paradigm--document-based distant supervision, which transfers the distantly supervised relation extraction as a document-level task and addresses it via document-level methods.
  Specifically, we first re-organize all sentences about the same head entity into an entity-centric document, rather than as a bag of sentences.
  Based on the document paradigm, we further transfer the distantly supervised relation extraction into the widely studied document-level MRC task.
  For example, as shown in Figure 1(b), all sentences about Obama are organized into a document, and then the birthplace of Obama is extracted by querying the document with questions such as ``Where was Obama born ?'';
  Finally, the relation tuple $\langle$\texttt{Obama}, \texttt{birthplace}, \texttt{United States}$\rangle$ will be extracted using the predicted answer ``United States''.
  Compared with the bag-based paradigm, the document-based distant supervision has the following advantages:
  1) By organizing all sentences about an entity into a document and extracting relations via querying the document with relation-specific questions, the document-based DS paradigm can effectively exploit all sentence-level, inter-sentence-level, and entity-level evidence;
  2) Because MRC models can be automatically trained to identify and detect the most relevant and plausible answers using $\langle$Document, Question, Answer$\rangle$ tuples, the document-based DS paradigm can inherently solve the noisy label problem;
  3) Our paradigm is easy to implement and very flexible, which can be easily extended by incorporating other information into documents. And more document-based models can be leveraged for the DSRE task, e.g., the Graph Convolutional Networks \cite{kipf_semi-supervised_2017}.
  
  Based on the document-based paradigm, this paper designs a BERT-based \cite{devlin_bert_2019} distantly supervised relation extraction method--DocDS, which extracts relations by querying documents with $\langle$head entity, relation, ?$\rangle$-style questions, and all candidate tail entities and a special NA entity (representing for ``no answer'') are used as candidate answers.
  For example, in Figure 1, the question ``Where was Obama born ?'' will be constructed for extracting $\langle$\texttt{Obama}, \texttt{birthplace}, ?$\rangle$.
  To train the DocDS effectively, we further propose a new loss function--DSLoss.
  By replacing the original cross-entry loss in MRC, DSLoss can effectively take two main problems of DS training into consideration: multi/noisy-answer problem (i.e., the same tail entity may be contained in multiple sentences, and some of them may be noisy) and the data imbalance problem( i.e., the wrong answers are far more than correct answers).
  We conduct experiments on the NYT dataset \cite{riedel_modeling_2010}, and experimental results show that our model achieves new state-of-the-art DS performance.
  
  The contributions of this paper are:
  \begin{enumerate}
  \item We propose a new DS paradigm--the document-based distant supervision for relation extraction, which can effectively leverage multi-level evidence and perform end-to-end denoising and extraction without any need for complicated components. And it is the first time to transfer the bag-based DSRE to the document-based paradigm to our best knowledge.
  \item Based on the above paradigm, we design a state-of-the-art DSRE method--DocDS, which formulates relation extraction as a document-based MRC task, and a new DSLoss is designed to train the model without answer annotations.
   \item Because our paradigm is easy to implement and very flexible, we believe it expands the DSRE research and can inspire more studies, e.g., more powerful document-based models, to address the DSRE problems.
  \end{enumerate}
  
  \section{Related Work}
  \textbf{DS for RE}. To resolve the labeled data bottleneck, \citet{mintz_distant_2009} first proposed distant supervision, which directly trains models on automatically generated training data.
  To resolve the noisy label problem \cite{riedel_modeling_2010,han_global_2016}, many studies adopt the multi-instance learning strategy \cite{riedel_modeling_2010, hoffmann_knowledgebased_2011,surdeanu_multiinstance_2012}, which represents all sentences containing the same entity pair as a bag, and then denoise them using special algorithms. These approaches are further extended by neural networks-based representation learning \cite{zeng_distant_2015,lin_neural_2016,feng_effective_2017,ji_distant_2017}, and the attention mechanism-based denoising \cite{lin_neural_2016,ji_distant_2017,du_multilevel_2018,han_global_2016, yuan_distant_2018, yuan_crossrelation_2018,jia_arnor_2019, ye_distant_2019, xing_distant_2019, chen_uncover_2019}.
  There are also some other techniques for resolving noisy label problem, e.g.,
  designing label adjustment/denoising strategies \cite{liu_softlabel_2017,luo_learning_2017,chen_uncover_2019},
  exploiting extra supervision resources \cite{lei_cooperative_2018,vashishth_reside_2018,wang_labelfree_2018,deng_leveraging_2019},
  selecting high-quality sentences via RL or adversarial training \cite{feng_reinforcement_2018,qin_dsgan_2018},
  applying structured learning method \cite{bai_structured_2019},
  alleviating noise with human participants \cite{zheng_diagnre_2019},
  and leveraging pre-trained language models \cite{alt_finetuning_2019}.
  
  \textbf{DS for NLP tasks}.
  Because labeled data bottleneck is a common challenge in NLP, distant supervision is also employed for many other NLP tasks, including named entity recognition (NER) \cite{yang_distantly_2018,ghaddar_robust_2018,peng_distantly_2019}, entity linking \cite{le_distant_2019}, entity-event extraction \cite{keith_identifying_2017}, code generation \cite{agashe_juice_2019}, open-domain question answering \cite{lin_denoising_2018}, etc.
  
  \textbf{Document-level RE}.
  Recently, document-level relation extraction has attracted increasing attention \cite{verga_simultaneously_2018,sahu_intersentence_2019,christopoulou_connecting_2019,nan_reasoning_2020}, and some large-scale document-level relation extraction datasets are also published \cite{yao_docred:_2019,wu_renet_2019}.
  Compared with these document-level RE methods:
  our DSRE method focuses on denoising and aggregating the multi-level evidence from multiple distantly labeled instances via constructing a pseudo document; meanwhile, the document RE focuses on spotting and reasoning of document information based on the structure of natural documents, e.g., multi-hop relation reasoning on discourse structure \cite{nan_reasoning_2020}.
  
  \textbf{MRC for NLP tasks}.
  Many MRC models have been proposed in recent years \cite{hermann_teaching_2015,seo_bidirectional_2016, cui_attentionoverattention_2017,chen_reading_2017}.
  Recently, due to its strong ability in information encoding and span spotting, MRC has been adopted as a basic technique for many NLP tasks, such as zero-shot relation extraction \cite{levy_zeroshot_2017a}, relation argument extraction \cite{roth_neural_2018}, knowledge graph building \cite{das_building_2018}, NER \cite{li_unified_2019}, etc.
  Compare with some studies directly leveraging MRC \cite{levy_zeroshot_2017a, das_building_2018}, we have a totally different main contribution, i.e., a new document paradigm which can provide multi-level evidence and perform denoising without any need for complicated components.
  
  \section{Document-based Distant Supervision via Machine Reading Comprehension}
  
  This section describes our document-based DS paradigm. Specifically, instead of representing sentences containing the same entity pair as a bag \cite{lin_neural_2016, ye_distant_2019}, we construct an entity-centric document for each head entity, so that all sentence-level, inter-sentence-level, and entity-level evidence about the same entity can be simultaneously exploited. Based on the document representation, DSRE can be effectively resolved via document-based methods, e.g., machine reading comprehension methods.
  
  In the following, we first introduce how to build entity-centric documents, then describe how to model relation extraction as a widely studied task--MRC, finally introduce the MRC model we used for relation extraction.
  
  \subsection{Entity-centric Document Building}
  
  In relation extraction, three types of information are considered useful \cite{zheng_aggregating_2016,vashishth_reside_2018, yuan_crossrelation_2018, ye_distant_2019}--sentence-level, inter-sentence-level and entity-level.
  The sentence-level information provides direct evidence about two entities (e.g., ``Obama was born in the United States'' directly expresses the birthplace relation between Obama and the United States).
  The inter-sentence-level information can provide evidence for more robust (if two sentences express the same relation) or more confident (if two sentences express two highly correlated relations, e.g., birthplace and nationality) relation prediction.
  For example, the birthplace instances ``Obama was born in the United States.'' can provide external evidence for the confidence of $\langle$\texttt{Obama}, \texttt{nationality}, \texttt{United States}$\rangle$. 
  The entity-level information can provide entity type evidence, which is useful for relation extraction because most relations have strong selectional preferences.
  For example, the head and tail entity of the birthplace must be person and location.
  
  To simultaneously exploit sentence-level, inter-sentence-level, and entity-level information for relation extraction, our document-based paradigm organizes all sentences about an entity as a pseudo document.
  Given a corpus, we first group all sentences containing the same entity pair and rank the sentences in each group in ascending order of length.
  After that, the groups of the same head entity are concatenated into a pseudo document.
  For example, in Figure 1(b), a document about Obama will be built by concatenating sentences about $\langle$\texttt{Obama}, \texttt{United States}$\rangle$, $\langle$\texttt{Obama}, \texttt{Hawaii}$\rangle$, etc.
  Because some head entities may have too many groups, and some groups may contain too many sentences, we limit each document containing up-to-$M$ tokens for efficiency (large documents will be divided into multiple smaller ones), and for each group, we only select the top-ranked $N$ sentences to add into the document. In this paper, $M$ and $N$ are empirically set to 300 and 15, respectively.
  
  \begin{table*}[!tp]
    \small
    \setlength{\belowcaptionskip}{-1em}
    \centering
      \begin{tabular}{C{8cm}C{5.5cm}}
      \toprule
      \multicolumn{1}{c}{\textbf{Relation}} & \multicolumn{1}{c}{\textbf{Question Pattern}} \\
      \midrule
      \texttt{business/company/founders} & \emph{Who is the founder of [Entity] ?} \\
      \texttt{location/country/languages\_spoken}  & \emph{Which language is spoken in [Entity] ?} \\
      \texttt{people/person/place\_of\_birth} & \emph{Where was [Entity] born ?} \\
      \texttt{location/location/contains} & \emph{[Entity] contains which place ?} \\
      \texttt{people/person/children} & \emph{Who is [Entity] 's child ?} \\
      \bottomrule
      \end{tabular}%
      \caption{Question patterns for several relations in the NYT dataset, where ``\emph{[Entity]}'' is the slot for head entity.}
    \label{tab:question}%
  \end{table*}%
  
  Based on the above document representation, the rich inter-sentence/entity information can be effectively leveraged via document-level models such as MRC. 
  For example, a RE model can effectively predict the birthplace of Obama by collecting together all related evidence, including the sentences expressing his birth, the sentences expressing his education, and his other relationships with the United States.
  
  \subsection{Relation-specific MRC Task Construction}
  
  Based on the entity-centric document representation, this section describes how to address DSRE by transferring it into a widely-studied document-level task–MRC.
  
  Specifically, given the constructed document about a head entity, we transfer relation extraction into an MRC task as: finding the tail entity (\bm{$t$}) that can answer the relation(\bm{$r$})-specific questions about the head entity (\bm{$h$}).
  Compared with previous methods \cite{lin_neural_2016, ye_distant_2019}, this schema aims to fill in the missing tail entity in $\langle h, r, ?\rangle$, rather than fill in the missing relation $\langle h, ?, t\rangle$ in traditional methods.
  We employ this schema because:
  1) It is easier and more natural to ask and answer $\langle h, r, ?\rangle$-style questions than $\langle h, ?, t\rangle$-style questions for MRC models.
  For example, it is easier for MRC models to answer ``Where was Obama born ?'' than ``What is the relation between Obama and the United States ?'', because the answer to the former question explicitly exists in the document (i.e., the tail entities), while the answer to the latter usually does not.
  2) The relation types about an entity are usually far fewer than the entity pairs in a dataset, therefore asking $\langle h, r, ?\rangle$-style questions is more efficient than asking $\langle h, ?, t\rangle$-style questions.
  
  Formally, to extract relation fact $\langle h, r, ?\rangle$ about the head entity $h$, we formulate the MRC task as a tuple $\{ \langle d_h, q_h^r, C\rangle, A\}$ (Figure 1(b) shows an example), where:
  \begin{itemize}
    \setlength{\itemsep}{0pt}
    \setlength{\parsep}{0pt}
    \setlength{\parskip}{0pt}
    \item $d_h=\{w_1, ...,w_i,..., w_{|d_h|}\}$ is the \textbf{pseudo document} of $e$, where $w_i$ is the $i$-th token.
    \item $q_h^r$ is the \textbf{relation-specific question} about $\langle{h, r, ?}\rangle$, e.g., ``Where was Obama born ?'' for $\langle$\texttt{Obama}, \texttt{birthplace}, ?$\rangle$.
    \item $C=\{c_1, ..., c_{|C|}\}$ are \textbf{tail entity candidates} in $d_h$ with an additional \emph{NA} entity for non-answer questions\footnote{Notice that MRC models can automatically detect answer spans in a document without identifying answer candidates. This paper identifies answer candidates to reduce the search space and improve the performance.}.
    Since each candidate entity appears multiple times in $d_h$, we represent $c_i$ as $\{c_{i,1},...,c_{i,j},...\}$, where $c_{i,j}$ is the $j$-th mention span of $c_i$. For example, in Figure 1(b), $\{$\emph{NA}, Hawaii, United States, Michelle$\}$ are candidate entities for $\langle$\texttt{Obama}, \texttt{birthplace}, ?$\rangle$ and the United States has two mention spans.
    \item $A=\{a_1, ..., a_{|A|}\}$ are the \textbf{correct answers} (distantly labeled tail entities holding relation $r$ with $h$) of the above question, and $A \subseteq C$.
  \end{itemize}
  
  \textbf{Relation-specific question construction}.
  Questions are critical for accurately extracting relations from the pseudo document.
  This paper constructs relation-specific questions by manually designing question patterns for each relation type.
  Table \ref{tab:question} shows several question patterns for relation types in NYT.
  For each head entity, we only ask questions consistent with its entity type. For example, we will not ask founder-of questions about Obama. 
  
  This paper uses one question pattern for a relation type because it is enough to achieve good performance. But our method can be easily extended by asking more questions.
  
  \subsection{MRC Model for Document-based Relation Extraction}
  Given $\langle d_h, q_h^r, C\rangle$ (i.e., the pseudo document, the relation-specific question, and the candidate entities) as input, an MRC model will assign each candidate entity a score, and the top-1 entity $c^*$ will be used as the final answer and form a relation tuple $\langle h, r, c^*\rangle$.
  Because our approach does not pre-assume any specific MRC models,
  we use the BERT-based MRC model \cite{devlin_bert_2019} as our base model.
  
  Specifically,  for the BERT-based MRC model, the input $x$ is the concatenated tokens of question $q_h^r$ and document context $d_h$ as:
  
  \noindent $\centering \ \ \ \ \ x = \{[\text{CLS}]..., q_i,..., [\text{SEP}], ..., w_i,..., [\text{SEP}]\}$
  
  \noindent with a special start token [CLS] and a separator token [SEP].
  For token representation, we sum up three standard embeddings--token, segment, and position embeddings used in \citet{devlin_bert_2019}, and additional indicator embedding for whether a token is in a candidate mention\footnote{We use pre-trained token, segment, and position embeddings provided by BERT, and randomly initialize the parameters of indicating embeddings and learn them during training.}.
  A multi-layer Transformer-based \cite{vaswani_attention_2017} encoder takes token representations as input and outputs their hidden representations--an $|x| \times d$-dimension matrix $\bm{{\rm H}}$, where each row refers to a token.
  An MLP layer is used to compute scores of each token being the start position and the end position of a correct answer:
  \begin{equation}
    \setlength{\abovedisplayskip}{4pt}
    \setlength{\belowdisplayskip}{4pt}
    [S, E] = \bm{{\rm H}} \cdot \bm{{\rm M}} 
  \end{equation}
  \noindent where $S$ is an $|x|$-dimension score vector indicating the score of each token in $x$ being the answer's start position, 
  $E$ is the same but indicating the scores of the end positions, $\bm{{\rm M}}$ is a $d \times 2$ parameter matrix.
  
  Based on the above scores, we estimate the probability of candidate entity $c_i$ as follows:
  \begin{align}
    \setlength{\abovedisplayskip}{0pt}
    \setlength{\belowdisplayskip}{-1pt}
    p(c_{i}|x) &= \sum_j p(c_{i,j}|x)\label{eq:output}\\
    p(c_{i,j}|x) &= \frac{e^{\textit{score}(i,j)}}{\sum_{\langle i',j'\rangle} e^{\textit{score}(i',j')}} \label{eq:span_output}\\
    \textit{score}(i,j) &= S[c_{i,j}^{\text{start}}] + E[c_{i,j}^{\text{end}}]\label{eq:score}
  \end{align}
  \noindent where $p(c_i|x)$ is the probability of entity $c_i$, $c_{i,j}$ is the $j$-th mention span of $c_i$, $c_{i,j}^{\text{start}}$/$c_{i,j}^{\text{end}}$ is the start/end position of $c_{i,j}$, and $\textit{score}(i,j)$ is the score of $c_{i,j}$.
  Besides, since the special NA entity does not exist in the document, we use the [CLS] as its mention, whose start and end positions are both 0.
  
  \textbf{Answer probability calibration with independent answer confidence.}
  Because a question may have multiple answers, and the answer probabilities in Eq. \ref{eq:output} are normalized on all tail entities of the same $\langle h, r, ?\rangle$ query, we need to calibrate the probability.
  For example, there are two answers to the question ``Who is the child of Obama ?'', therefore at most one answer can be assigned probability greater than 0.5 in Eq. \ref{eq:output}.
  Specifically, we calibrate answer probabilities by computing the independent answer confidence:
  \begin{equation}
    \setlength{\abovedisplayskip}{4pt}
    \setlength{\belowdisplayskip}{4pt}
    \begin{aligned}
    \text{conf}(c_i|x) &= \frac{p(c_{i}|x)}{p(c_{i}|x)+ p(\text{NA}|x)}
    \end{aligned}
  \label{eq:conf_calibration}
  \end{equation}
  \noindent where $p(\text{NA}|x)$ is the non-answer probability and $\text{conf}(c_i|x)$ is now independent with other candidate tail entities.
  Notice that Eq. \ref{eq:conf_calibration} is only used as confidence score, and for model learning we still use $p(c_i|x)$ in Eq. \ref{eq:output}.
  
  \section{DSLoss for Model Learning}
  The standard MRC models are trained using the cross-entropy loss:
  \begin{equation}
    \setlength{\abovedisplayskip}{5pt}
    \setlength{\belowdisplayskip}{5pt}
    {\rm CELoss}(x,y) = \sum_{\{x, y\}} - \log p(y|x)
  \end{equation}
  \noindent where $\{x, y\}$ is a data sample with input $x=\langle \text{document}, \text{question} \rangle$ and its label $y$.
  The cross-entropy loss, however, cannot be directly used to train our models due to two problems:
  
  1). \textbf{Multi/noisy-answer problem}. In document-based DS, a query about $\langle$document, question$\rangle$ is only labeled with entity answer, rather than span answer. Because an entity may have multiple mention spans, we do not know which of them are correct labels, and which of them are noisy labels. 
  
  2). \textbf{Data imbalance problem}. In the classical MRC task, most questions have answers (positive samples). However,  in our MRC-based relation extraction, many relation-specific questions do not have answers (negative samples). The model will over-fit to negative samples if we directly optimize the cross-entropy loss.
  
  To resolve the above problems, we design a new loss function, named \textbf{DSLoss} (distant supervision loss), which uses a noise-tolerant loss function $L_n(x, y)$ to address multi/noisy-answer problem and a risk-sensitive factor $r(x)$ to balance between positive/negative samples:
  \begin{equation}
    {\rm DSLoss}(x, y) = r(x) \cdot
    L_n(x, y)
  \end{equation}
  
  \textbf{Noise-tolerant Loss Function}.
  Given a training instance $\{\langle d_h, q_h^r, C\rangle, A\}$, we denote the input as $x=\langle d_h, q_h^r, C\rangle$, target as $y=A$. The noise-tolerant loss function is:
  
  \begin{align}
    \setlength{\abovedisplayskip}{-4em}
    \begin{split}
       L_n(x, y) = &- \sum_{a_i \in A} \sum_{a_{i,j}}w_{ij}\log p(a_{i,j}|x) \\
      &+\lambda\sum_{o \in\ C-A} p(o|x) \log p(o|x)
    \end{split}\\
    w_{ij} =& \frac{p(a_{i,j}|x)}{\sum_{j'}p(a_{i,j'}|x)}
    \label{eq:noise-tolerant}
  \end{align}
  \noindent where $a_i$ is a correct answer in $A$, $a_{i,j}$ is its $j$-th mention span, $p(a_{i,j}|x)$ is the probability by Eq. \ref{eq:span_output}, $w_{ij}$ is a confidence weight of the $j$-th span over all spans of $a_i$, and we have $w_{ij} = 1$.
  The second term is a regularization term, where $o$ is a wrong non-NA answer, $p(o|x)$ is the probability by Eq. \ref{eq:output}. $\lambda$ is a hyper-parameter and is set to 0.1 in our experiment.
  
  The noise-tolerant loss function can effectively resolve the multi/noisy-answer problem because:
  1) It can implicitly select one best plausible answer using the confidence weight $w_{ij}$.
  Because the $L_n$ reaches its minimization only if for each correct entity, only one mention span's confidence weight approaches 1 while the others are near 0.
  As a result, the model will implicitly assign the highest probability to the most confident span for each answer entity.
  2) It can learn to maximize the probability margin between correct entities and other entities using the second term, as explained in \citet{lin_cost_2019}.
  The max-margin is a common strategy for multi-label learning \cite{taskar_max_2004, ye_deep_2019}, and the noise-tolerant loss function can implicitly incorporate it.
  
  \textbf{Risk-sensitive Factor}.
  To address the data imbalance problem, an effective solution is assigning cost-sensitive weights to different instances. Inspired by the Focal loss \cite{lin_focal_2018} which assigns higher weights to hard positive samples, we balance instances using a risk-sensitive factor:
  \begin{equation}
    r(x) = (1 - \max_{a \in A} p(a|x) + \max_{o \in P} p(o |x))^\gamma
  \end{equation}
  \noindent where $P$ is the set of all entities except for the NA entity (this paper empirically tunes $\gamma$ as 2).
  
  Using the risk-sensitive factor, DSLoss resolve the data imbalance problem by assigning higher weights to hard and positive samples than easy negative ones:
  1) For the hard samples whose predictions are not correct, their risk factors are $\geq$ 1.0;
  2) For the easy samples whose predictions are correct, the risk factors can balance between the positive samples (whose weights are always 1.0) and negative samples (whose weights are always less than 1.0).
  
  \section{Experiments}
  
  \subsection{Experimental Settings}
  
  \begin{figure*}[!htp]
    \begin{minipage}[b]{0.95\columnwidth}
      \setlength{\abovecaptionskip}{-0.1em}
      \setlength{\belowcaptionskip}{-0.5em}
      \centering
      \includegraphics[width=\columnwidth]{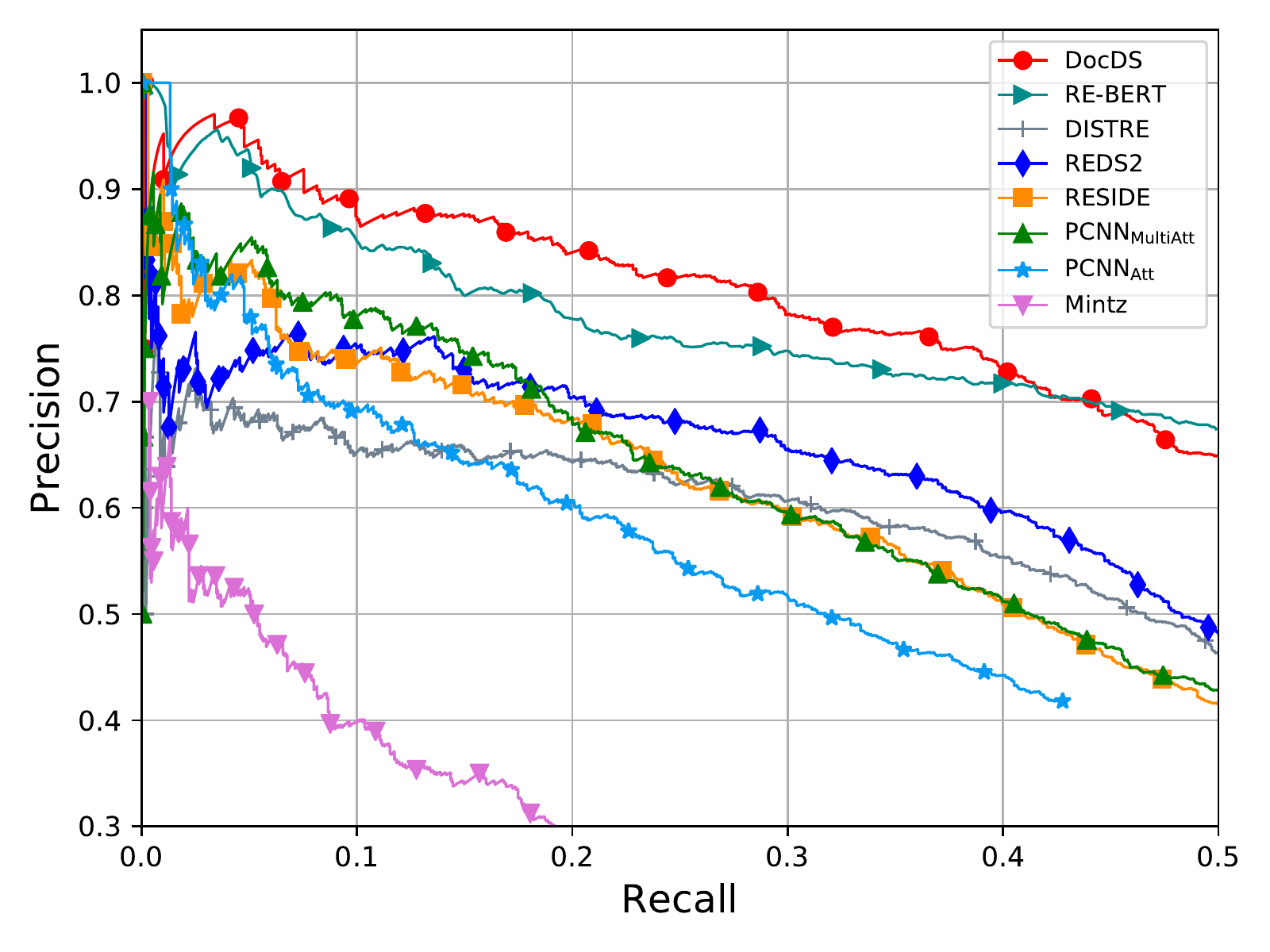}
      \caption{Precision-recall curves of all baselines and our method.}
      \label{fig:held-out}
    \end{minipage}
    \hspace{2em}
    \begin{minipage}[b]{0.95\columnwidth}
      \setlength{\abovecaptionskip}{-0.1em}
      \setlength{\belowcaptionskip}{-0.5em}
      \centering  
      \includegraphics[width=\linewidth]{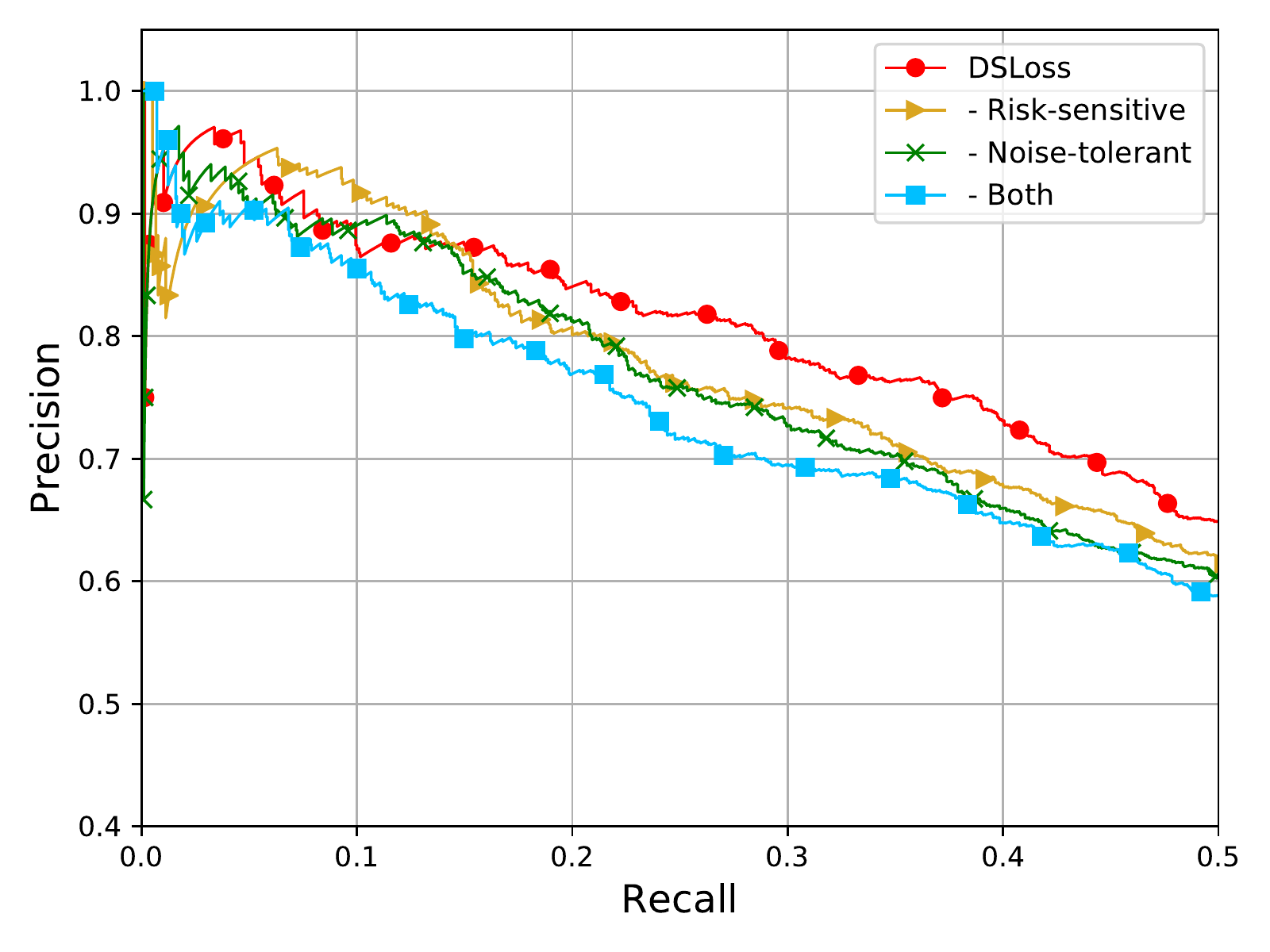}
      \caption{Precision-recall curves of our model using different loss functions.}
      \label{fig:ablation}
    \end{minipage}
  \end{figure*}
  
  \textbf{Dataset}.
  Following previous studies \cite{riedel_modeling_2010,lin_neural_2016}, this paper uses the New York Times (NYT) dataset \cite{riedel_modeling_2010}. NYT is constructed by distantly aligning relations in Freebase with sentences from the New York Times corpus \cite{bollacker_freebase_2008}, and the data of years 2005-2006 is used for training and 2007 for testing.
  The training data contains 522,611 sentences and 281,270 entity pairs.
  The test data contains 172,448 sentences and 96,678 entity pairs.
  We further randomly sample 20\% training data for validation and leave the remaining for training.
  
  \textbf{Baselines}.
  We denote our method as DocDS and compare its performance with the following four types of baselines:
  
  1). \textbf{Naive DS} method--\emph{Mintz}\cite{mintz_distant_2009} which uses feature-based classifiers.
  
  2). \textbf{Bag-based neural networks}, including \emph{PCNN$_\text{Att}$}\cite{lin_neural_2016} which denoises sentences using an attention mechanism, and \emph{PCNN$_\text{MultiAtt}$}\cite{ye_distant_2019} which further uses inter-bag attentions to resolve bag-level noise.
  
  3). \textbf{Bag-based neural networks with external knowledge}, including \emph{RESIDE}\cite{vashishth_reside_2018} utilizing additional side information from KBs, and \emph{REDS2} \cite{deng_leveraging_2019} leveraging relational tables from the Web.
  
  4). \textbf{Bag-based neural networks with pre-trained language models (PLMs)}, including \emph{DISTRE} \cite{alt_finetuning_2019} using the pre-trained GPT model, and \emph{RE-BERT}\cite{wu_practical_2020} using the pre-trained BERT model with a transitional loss.
  
  \begin{table}[!t]
    \setlength{\belowcaptionskip}{-1em}
    \centering
      \begin{tabular}{cc}
      \toprule
      \textbf{Hyper-parameter} & \textbf{Value} \\
      \midrule
      Weight Initialization & BERT$_\text{base}$ \\
      Learning Rate & 3e-5(with Adam) \\
      Warmup Proportion  & 0.1 \\
      Batch Size & 32 \\
      Dropout Rate & 0.1 \\
      Max Sequence Length & 384 \\
      \bottomrule
      \end{tabular}%
      \caption{Main hyper-parameter settings.}
    \label{tab:parameter}%
  \end{table}%
  
  \textbf{Evaluation Criteria}.
  Following previous studies \cite{lin_neural_2016,deng_leveraging_2019}, we evaluate all methods by directly comparing the extracted relational tuples with those in the Freebase.
  And the precision-recall curve, the area under the curve (AUC), and Precision@N (P@N) values are used to assess different methods.
  
  \textbf{Other Settings}.
  Table \ref{tab:parameter} shows the main hyper-parameters.
  All experiments rely on the PyTorch implement of BERT \cite{wolf_huggingfaces_2019}.
  And we train each model on a single Nvidia TiTan RTX\footnote{Our code is available at https://www.github.com/lingyongyan/docds.}.
  
  \subsection{Overall Performance}
  
  \begin{table*}[!t]
    \setlength{\belowcaptionskip}{-1em}
    \centering
    %\resizebox{\textwidth}{!}{
    \begin{tabular}{C{5em}cccccc}
      \toprule
      \multicolumn{2}{c}{\textbf{Method}} & \textbf{AUC} & \textbf{P@100} & \textbf{P@200} & \textbf{P@300} & \textbf{Mean} \\
      \midrule
      \multicolumn{1}{C{5em}}{Naive} & \multicolumn{1}{l}{Mintz} & 0.107 & 0.523 & 0.502 & 0.450 & 0.492\\
      \midrule
      \multicolumn{1}{C{5em}}{\multirow{2}[0]{*}{\tabincell{c}{Bag Based}}} & \multicolumn{1}{l}{PCNN$_\text{Att}$} & 0.342 & 0.730 & 0.680 & 0.673 & 0.694\\
            & \multicolumn{1}{l}{PCNN$_\text{MultiAtt}$} & 0.422 & 0.918 & 0.840 & 0.787 & 0.848\\
      \midrule
      \multicolumn{1}{C{5em}}{\multirow{2}[1]{*}{\tabincell{c}{\hspace{-0.4em}Bag Based\\\hspace{-0.4em}\small{(with knowledge)}}}} & \multicolumn{1}{l}{RESIDE} & 0.415 & 0.818 & 0.754 & 0.743 & 0.772\\
            & \multicolumn{1}{l}{REDS2} & 0.447 & 0.824 & 0.796 & 0.766 & 0.795 \\
      \midrule
      \multicolumn{1}{C{5em}}{\multirow{2}[1]{*}{\tabincell{c}{\hspace{-0.4em}Bag Based\\\hspace{-0.4em}\small{(with PLMs)}}}} & \multicolumn{1}{l}{DISTRE} & 0.422 & 0.680 & 0.670 & 0.653 & 0.668\\
      & \multicolumn{1}{l}{RE-BERT*} & - & 0.920 & 0.860 & 0.823 & 0.868\\
  \midrule
      \multicolumn{1}{C{5em}}{\multirow{2}[1]{*}{\tabincell{c}{Doc Based}}} & \multicolumn{1}{l}{Bert-Two-Step} & 0.443 & 0.820 & 0.755 & 0.717 & 0.764 \\
       & \multicolumn{1}{l}{\textbf{DocDS}} & \textbf{0.595} & \textbf{0.939} & \textbf{0.889} & \textbf{0.873} & \textbf{0.900}\\
      \bottomrule
    \end{tabular}%
    %}
    \caption{P@N and AUC scores of all baselines and our method. \emph{* The AUC score is not report in RE-BERT.}}
    \label{tab:held-out}%
  \end{table*}%
  
  Figure \ref{fig:held-out} and Table \ref{tab:held-out} show the overall performance of our method and all baselines on the held-out test data.
  We can see that:
  
  1). \textbf{The document-based distant supervision provides a new and effective DS paradigm}. On NYT, our method achieves the new state-of-the-art performance (the best AUC score--0.595). Particularly, compared with most baselines,  our model can achieve at least 33.1\% AUC improvement; compared with the BERT-based model \emph{RE-BERT}, our model can also achieve higher values on all P@N metrics. We believe this is because the document representation provides an effective way to exploit inter-sentence-level and entity-level evidence, and the noisy label problem can be inherently resolved using MRC models.
  
  2). \textbf{By exploiting sentence-level, inter-sentence-level, and entity-level evidence simultaneously, our method significantly outperforms bag-based methods}.
  In Table \ref{tab:held-out}, DocDS achieves around 74.0\% and 41.0\% AUC improvements over two bag-based baselines: \emph{PCNN$_\text{Att}$} and \emph{PCNN$_\text{MultiAtt}$}.
  Furthermore, even for most bag-based baselines which leverage external knowledge or pre-trained language models--\emph{RESIDE}, \emph{REDS2} and \emph{DISTRE}, our method can still significantly outperform them by at least 33.1\% AUC improvement.
  This verifies the effectiveness of further leveraging the sentence-level, inter-sentence-level, and entity-level information for relation extraction.
  
  3). \textbf{By training MRC models using $\langle$document, question, answer$\rangle$ tuples with DSLoss, our method can inherently solve the noisy label problem of distant supervision}. 
  The noisy label problem is challenging for DS methods: we can see that without taking this problem into consideration, \emph{Mintz} can only achieve a 0.107 AUC score.
  Compared with the bag-based DS baselines (\emph{PCNN$_\text{Att}$} and \emph{PCNN$_\text{MultiAtt}$}) and the bag-based DS baselines using external knowledge or language models (\emph{RESIDE}, \emph{REDS2}, \emph{DISTRE} and \emph{RE-BERT}), our method can achieve the best performance.
  This verifies the effectiveness of the document-based paradigm for resolving noisy label problem.
  
  \subsection{Detailed Analysis}
  This section analyzes our model in detail, including the effect of language models, the effect of document setting, the effect of the DSLoss, and the effect of sentence number.
  
  \textbf{Effect of Language Models}.
  To analyze the effect of language models, we conduct the following comparison experiments on different model settings: the two-step Bert-based document-level classifier \cite{wang_fine-tune_2019} upon the constructed pseudo document, denoted as \emph{Bert-Two-Step}; and our model (\emph{DocDS}).
  The results are shown in Table \ref{tab:held-out} and Figure \ref{fig:held-out}.
  And we can see that:
  \emph{Bert-Two-Step} can achieve competitive performance (0.443 AUC score) compared to most bag-based baselines, but its performance gain is still limited compared to \emph{DocDS} (0.595 AUC score).
  This may be mainly due to two reasons:
  1) Bert-based classifier cannot effectively deal with the noisy sentences in the constructed pseudo document, while MRC models can inherently address this problem because MRC models usually aim to detect valuable evidence for specific questions rather than aggregate noisy sentences.
  2) Bert-based classifier still suffers from the multi-label problem and the data imbalance problem, which is proved to impact the model performance in the following experiments.
  
  \begin{table}[!tp]
    \setlength{\belowcaptionskip}{-1em}
    \centering
      \begin{tabular}{p{4.5em}C{2em}C{4.4em}C{4.4em}}
      \toprule
      \multicolumn{1}{c}{\textbf{Method}}  &  \textbf{AUC}  &  \textbf{AUC(\small{Multi})}  &  \textbf{AUC(\small{Single})}  \\
      \midrule
      DocDS$_\text{sent}$  &   0.550    &    0.685   & 0.481 \\
       DocDS$_\text{pair}$  &   0.575    &    0.705   & 0.493 \\
       DocDS  & \textbf{0.595} &    \textbf{0.728}   & \textbf{0.510} \\
      \bottomrule
      \end{tabular}%
    \caption{Performance under different document settings, where DocDS$_\text{sent}$ builds a document only using one sentence for an entity pair, DocDS$_\text{pair}$ builds a document for each entity pair.}
    \label{tab:document_settings}%
  \end{table}%
  
  \textbf{Effect of Document}.
  To analyze the effects of different types of information, we conduct experiments on different document settings:
  for each entity pair, we randomly select only one sentence as its document--DocDS$_\text{sent}$;
  for each entity pair, we build a single document using all its sentences--DocDS$_\text{pair}$;
  the original document setting--DocDS.
  For investigation, we also report the AUC scores of different settings on two subsets of test data--Multi(the relational facts with more than one sentence) and Single (the relational facts with only one sentence).
  
  Table \ref{tab:document_settings} shows the results on different document settings, and we can see that:
  1) It is effective to take sentence-level, inter-sentence-level and entity-level evidence together for relation extraction – the DocDS can significantly outperform DocDS$_\text{sent}$ and DocDS$_\text{pair}$.
  2) MRC models can effectively exploit evidence for multiple sentences--both DocDS and DocDS$_\text{pair}$ obtain a significant performance improvement on Multi subset than on Single subset (from 0.510 to 0.728 for DocDS and 0.493 to 0.705 for DocDS$_\text{pair}$.
  3) Both inter-sentence-level and entity-level evidence are helpful for relation extraction--By further exploiting inter-sentence evidence, DocDS$_\text{pair}$ can achieve 4.5\% AUC improvement than DocDS$_\text{sent}$; and by further exploiting entity-level evidence, DocDS further achieves 3.5\% AUC improvement than DocDS$_\text{pair}$.
  The above results verify both the effectiveness of the document-based representation and of the document-based MRC model for exploiting different evidence.
  
  \begin{table}[!tb]
    \setlength{\belowcaptionskip}{-1em}
    \centering
      \begin{tabular}{L{7em}cc}
      \toprule
      \multicolumn{1}{c}{\textbf{Loss Function}} & \textbf{AUC} & $\bm{\Delta}$ \\
      \midrule
      \ \ DSLoss & \textbf{0.595} & - \\
      \midrule
      \ \ - Risk-sensitive & 0.566 & -4.9\%\\
      \ \ - Noise-tolerant & 0.563 & -5.4\%\\
      \ \ - Both & 0.543 & -8.7\%\\
      \bottomrule
      \end{tabular}%
      \caption{Performance of different loss functions.}
    \label{tab:ablation}%
  \end{table}%
  
  \textbf{Effect of DSLoss}.
  The DSLoss plays a central role in resolving the noisy label problem in DS.
  To analyze its effect, we conduct an ablation study by ablating the risk-sensitive part (-risk sensitive) and the noise-tolerant part (-noise tolerant), i.e., replacing the noise-tolerant loss with the classical cross-entropy loss, which uniformly enlarges the likelihoods of all matched entity mentions.
  
  The performance is shown in Figure \ref{fig:ablation} and Table \ref{tab:ablation}.
  We can see that:
  1) our DSLoss is effective for training document-based DS methods;
  2) due to the multi/noisy-answer problem and the data imbalance problem, both the risk-sensitive part and the noise-tolerant part are useful: the performance will decrease when ablating any of them (-4.9\% AUC score by ablating the risk-sensitive part and -5.4\% AUC score by ablating noise-tolerant part).
  We can also see that in Figure \ref{fig:ablation}, the precision drops more quickly than others when the recall increases if we do not use either part of the DSLoss.
  
  \begin{table}[!tbp]
    \setlength{\belowcaptionskip}{-1em}
    \centering
      \begin{tabular}{ccc}
      \toprule
      \# of sentences & AUC(val) & AUC(test) \\
      \midrule
      5     & 0.767 &  0.577 \\
      10    & 0.761 & 0.571 \\
      15    & \textbf{0.771}  & \textbf{0.595} \\
      20    & 0.767 &  0.584 \\
      \bottomrule
      \end{tabular}%
    \caption{Performance of different sentence numbers, where AUC(val) is the AUC score on the validation data, and AUC(test) is the AUC score on the test data.}
    \label{tab:sentence}%
  \end{table}%
  
  \textbf{Effect of Sentence Number}.
  As we described in Section 3, we only use top $N$ sentences for each entity pair, in order to avoid too large document.
  To evaluate the influence of sentence numbers,
  we conduct experiments with different sentence numbers, and their results are shown in Table \ref{tab:sentence}.
  From Table \ref{tab:sentence}, we can see that:
  1) Our model benefits from leveraging more sentences as evidence (The AUC score increase from 0.577 to 0.595 by increasing the sentence number from 5 to 15).
  2) Our model achieves its best performance on both validation and test data using a maximum sentence number 15.
  3) Our model is robust to the number of sentences: fewer or more sentences will only result in a slight performance variation.
  This may be because there is a trade-off between useful and noisy information:
  fewer sentences (e.g., 5) may miss some evidence; more sentences (e.g., 20) may introduce more noisy information, although they will contain more useful evidence.
  
  \section{Conclusion}
  This paper proposes a new DS paradigm--the document-based distant supervision, which can effectively exploit all sentence-level, inter-sentence-level, and entity-level evidence for relation extraction, and inherently resolve the noisy label problem by modeling relation extraction as the MRC task. Specifically, a document-based DS method--DocDS is proposed to predict relational facts, and a new loss function--DSLoss is designed for effective and balanced model learning. Experimental results show that the document-based DS paradigm can achieve state-of-the-art performance. For future work, we want to design more document-based DS approaches, and because DS is widely used in NLP, we also want to apply our method to other tasks such as NER and event extraction.

\bibliographystyle{acl_natbib}
\bibliography{dsre_reference}

\end{document}